\documentclass[12pt]{interact}

\usepackage[utf8]{inputenc}
\usepackage{hyperref}
\usepackage{float}
\usepackage{multirow}
\usepackage[parfill]{parskip}
\usepackage{xcolor}
\usepackage{graphicx}
\usepackage{array}
\usepackage{comment}
\usepackage{booktabs}
\usepackage{colortbl}
\usepackage{multicol}
\usepackage{pgfplots,siunitx}
 \usepackage{setspace}

\usepackage{xcolor}
\definecolor{UmUBlue}{RGB}{42,71,101}
\definecolor{UmUGreen}{RGB}{115,167,144}
\definecolor{UmUGold}{RGB}{215,177,124}
\definecolor{UmUPink}{RGB}{234,186,185}
\definecolor{NoUmUColour}{RGB}{84,142,202}
 
\usepackage{tikz}
\usetikzlibrary{positioning}
\usetikzlibrary{calc,backgrounds,fit}
\usetikzlibrary{shapes.geometric}

\usepackage[caption=false]{subfig}

\usepackage{natbib}

\pgfplotsset{compat=1.17}

\usepackage{times}

\title{AI-Driven Contextual Advertising:\\[.5ex] A Technology Report and Implication Analysis}

\definecolor{lightblue}{rgb}{0.3,0.4,0.6}

\begin{document}

\textbf{\large AI-Driven Contextual Advertising:\\ A Technology Report and Implication Analysis}\\[1ex]

{\large Emil Häglund and Johanna Björklund}

{\it Department of Computing Science, Umeå University, Umeå, Sweden}

Department of Computing Science\\ 
Umeå University, 901 87 Umeå\\ 
Sweden\\[1ex]
\texttt{emilh@cs.umu.se}\\[1ex]
+46 73-501 03 17

\vspace{5ex}

\textbf{Emil Häglund} is a PhD Student at the Department of Computing Science at Umeå University. His research is on how artificial intelligence (AI) can be applied to online advertising and the consequences for citizens and society when AI becomes the dominating method of placing messages in online media contexts. Emil received his MSc in Computer Science at KTH Royal Institute of Technology in 2020. 

\textbf{Johanna Björklund}  is an Associate Professor at the Department of Computing Science at Umeå University. Her research is on semantic, or human-like, analysis of multimodal data, incorporating, e.g., images, audio, video, and text. She is also a co-founder of the media tech company Codemill which provides cloud-based tools for video production to, e.g., ITV, BBC, and ProSieben, and a co-founder of Adlede, which provides contextual advertising solutions to, e.g., IKEA and American Express.  Johanna received her PhD in Computer Science at Umeå University in 2007. After her dissertation, she worked for a period of time at the Technical University of Dresden as a research assistant at the chair of Prof. Vogler, before returning to Umeå and a position as first a junior and later a senior lecturer, becoming a Docent in 2016. Her work is supported by the Swedish Research Council, the Swedish Foundation for Strategic Research, Knut and Alice Wallenberg's Foundation, and various EC funding programs. She is the director of the Wallenberg Research Arena for Media and Language, which part of the research program Wallenberg AI, Autonomous Systems, and Software Program.

\clearpage 
\doublespacing

\setlength{\parindent}{4em}
\setlength{\parskip}{0em}

\maketitle

\begin{abstract}
Programmatic advertising consists in automated auctioning of digital ad space. Every time a user requests a web page, placeholders on the page are populated with ads from the highest-bidding advertisers. The bids are typically based on information about the user, and to an increasing extent, on information about the surrounding media context. 
The growing interest in contextual advertising is in part a counterreaction to the current dependency on personal data, which is problematic from legal and ethical standpoints. The transition is further accelerated by  developments in Artificial Intelligence (AI), which allow for a deeper semantic understanding of context and, by extension, more effective ad placement. In this article, we begin by identifying context factors that have been shown in previous research to positively influence how ads are received. We then continue to discuss applications of AI in contextual advertising, where it adds value by, e.g., extracting high-level information about media context and optimising bidding strategies. However, left unchecked, these new practices can lead to unfair ad delivery and manipulative use of context. We summarize these and other concerns for consumers, publishers and advertisers in an implication analysis.\\

\noindent
Contextual targeting; artificial intelligence; advertising effectiveness
\end{abstract}



\noindent 
Contextual advertising is the practice of optimizing advertising effectiveness by placing ads in favourable media contexts. Common objectives are to improve brand perception, increase click-through rate, or create purchase intent. There is a substantial literature on the impact of context on advertising effectiveness, but the interplay between ad and context is complex, and there are few certainties. Despite this, contextual advertising is increasingly used for online media, where data about the media context informs bidders in automated auctions of digital ad space. The practice is already established in search advertising, to include sponsored links among the answers to search queries. It is now becoming a commodity also for display advertising, that is, the promotional third-party banners that finance a large part of the world-wide web.

The increasing use of contextual advertising can be ascribed to two factors. First, the concern for consumer privacy and fairness which led to General Data Protection Regulation (GDPR) and the California Consumer Privacy Act (CCPA). Furthermore, tech giants are increasingly reducing targeting capabilities, most notably Google plans to remove third-party cookies from Chrome in 2023 \citep{Reuters:2021}. This motivates the industry to move away from personal data and behavioural tracking~\citep{IAB:2021}, making contextual advertising, which requires neither, an attractive alternative. 
Rather than looking at previously recorded data, contextual advertising assumes that consumers' content requests are indicative of their current state of mind and product preferences, and places ads accordingly. 

A second driver is the abundance of data in an online environment, which makes it possible  to harness recent developments in artificial intelligence (AI) for targeting purposes. In fact, with billions of ads being served on a daily basis, automation is a necessity. Moreover, in contrast to the case for, e.g., autonomous vehicles and medical diagnostics, the consequences of mis-classifications are generally acceptable. This facilitates experimentation and lowers the barriers for AI applications in online advertising. In the case of contextual advertising, modern AI technology can analyse media content beyond the superficial level of keywords, and provide a deeper and more refined profiling of contexts. It can detect signals like topic, sentiment, visual complexity or imagery, target particular messages, and optimize bidding in programmatic real-time auctions for advertising space. While AI is already being applied to contextual advertising, it is still at an early stage of adoption. 

Although contextual advertising avoids issues related to the use of personal data, there are nonetheless risks and concerns linked to contextual advertising. These risks can be amplified by AI and automation. Content preferences can be used as a proxy for demography, and lead to discrimination against different communities. An automated, semi-autonomous, ad distribution system could find that targeting specific groups is beneficial when aiming to maximize clicks. While skewed ad delivery is a potentially serious outcome, it can be difficult to discover or predict due to the lack of interpretability in many AI systems. 
Finally, if contextual advertising is combined with personalized advertising, the advertiser knows not only who the reader is, but also what he or she is reading about, resulting in  severe privacy risks.


In this article, we first cover previous work on contextual advertising and extract a list of commonly studied context factors. By examining the influence of these context factors on the viewer's perception of an ad, i.e., their priming effects, we can better understand what opportunities there are for automation.  We continue in this line with a discussion of the application of AI to contextual advertising, detailing how AI technology is used to target different types of media contexts. We conclude by discussing the implications of AI-driven contextual advertising, raising issues of concern for consumers, advertisers and media publishers. 



\section{Related work}
Advertisers have long realized how the right context can direct attention towards an ad, or increase the persuasiveness of the ad message. For example, by placing ads for public transport on a billboard next to a congested highway, the message reaches people facing a problem to which the ad provides a solution. To target customers with an interest in nature, a company selling outdoor products could pay to include a supplement advertising their products in a magazine about hunting and fishing. More generally, advertisers can use context to address groups of consumers with specific interests or shared challenges. When operating in an online setting, they can target contexts across the entire world wide web and so achieve global reach. 

Early research on the influence of context on advertising effectiveness mainly focused on print and TV media, while a large share of the more recent studies examine display advertising, the rectangular ads that pervade websites and social media. 
Advertising effectiveness is a multifaceted notion and can be evaluated in different ways. Previous studies have considered metrics such as attention,  recall and  purchase  intention.  Three underlying context factors have received particular attention: (1) the applicability of content and ad, (2) the affective tone of the content, and (3) the involvement of the consumer. We briefly summarise previous work on each of these in order.

\subsection{Applicability}

Priming describes the influence of context features on the consumer's cognitive and behavioural reactions to an ad. By inducing feelings or emphasizing attributes, the consumer is led towards certain associations, which in turn affects how the ad is perceived. In a study by \cite{Yi:1990}, subjects were shown an article together with a companion ad for a car, that emphasized the large size of the car. When the topic of the article was safety in air travel, the subjects tended to associate the size of the car with safety; when the topic was oil entrepreneurs, the subjects instead associated the size with poor fuel economy. The theme of the priming article influenced the subjects' attitude towards the brand and advertisement, as well as their purchase intention. 

The notion of applicability describes the degree of overlap between the contextual primes and the target stimuli. In the case of contextual advertising, applicability can be understood as a topical similarity between context and ad. Strong applicability has been found to increase consumer's attitude towards brands, and to improve their recall of the ad, in print media \citep{Shen:2007,Yi:1990}, TV \citep{Furnham:2002}, and display advertising \citep{Huang:2014,Song:2014}. \cite{Wojdynski:2016} complemented these studies by showing that news readers pay more attention to applicable ads in also in online contexts.


\subsection{Affective tone}

As a rule, advertisers want their brand and products to appear in positive contexts. The basis for this is affective priming, i.e., the observation that the affective tone of media content induces feelings that shape how ads are received. Studies have confirmed that positive content improves attitudes and purchase intention in both print media and television~\citep{DePelsmacker:2002,Goldberg:1987, Kamins:1991, Yi:1990}. However, a congruity in tone between ad and content (in other words, positive-positive or negative-negative) leads to similar effects \citep{DePelsmacker:2002,Kamins:1991}.

\subsection{Content involvement}

Priming theory suggests that when users appreciate and are engaged in media content, their perception of accompanying ads improve~\citep{Goldberg:1987}. However, if the users are so deeply emerged that they fail to notice the ads, the positive priming is lost. Advertisers employ a range of methods to draw the reader's attention towards the ads, for example, enhancing colors~\citep{Fernandez:2000} or positioning in the media~\citep{Huang:2018}, but these may spoil the user experience. There is a marked tension between stakeholder interests: The consumers want to optimise their user experience, while the advertisers want to achieve their marketing goals. The publishers need to find balance between the two, where both consumers and advertisers are sufficiently satisfied to continue their engagement with the publisher.  

The effects of content involvement on advertising effectiveness depends on the type of medium being consumed. Studies have found that TV viewers who are engaged in what they are watching, show increased ad recall and more positive attitude towards the ad served~\citep{DePelsmacker:2002, Tavassoli:1995}. Among readers of print media on the other hand, a higher degree of involvement \emph{reduces} ad recall \citep{Norris:1992, DePelsmacker:2002}. We conjecture that this difference can be explained by the mode in which ads are presented. In TV, ads temporarily replace the media stream, forcing themselves upon the consumer's attention, whereas in print media, an absorbing article is more likely to keep the consumer's attention away from  accompanying ads. 
Differences across media channels and the interplay of various context factors produces a complex environment with few certainties. The promise of AI within contextual advertising lies in disentangling this complexity, and identifying favourable ad placements using the richness of contextual data available online. 

\section{Artificial intelligence}
\label{sec:ai}
The volume and speed at which online advertising space is bought and sold, in combination with the economic values at stake, have motivated immense investments in process automation. This has made online advertising one of the early success stories for applied AI~\citep{QinJiang:2019}. Initial use has focused on bidding strategies and personalization, but we  expect that AI will come to play a prominent role also in contextual advertising, where recent advances in media AI can be leveraged to match contexts with ads. The same technologies that make it possible to automatically caption images with text, or realize visual-linguistic question answering, can help uncover the dynamics between ads and context (see, for example, \citep{Chen:2016,Sujuan:2017}), and so improve advertising effectiveness. Initial solutions have already been marketed by leading advertising platforms and the technology is gradually becoming more sophisticated.

The vast majority of online ads are traded in programmatic auctions~\citep{Forbes:2018,Busch:2016}, where  advertising space is sold by digital publishers to advertisers.  Since the auctions are conducted at the basis of individual viewings, they need to be completed within matters of milliseconds, and the number of daily auctions is counted in tens of billions. Moreover, the value of a won auction is revealed immediately to the buyer: either the ad is clicked, or it is not. The combination of speed, scale, and direct feedback make programmatic auctions a prime application area for AI. 

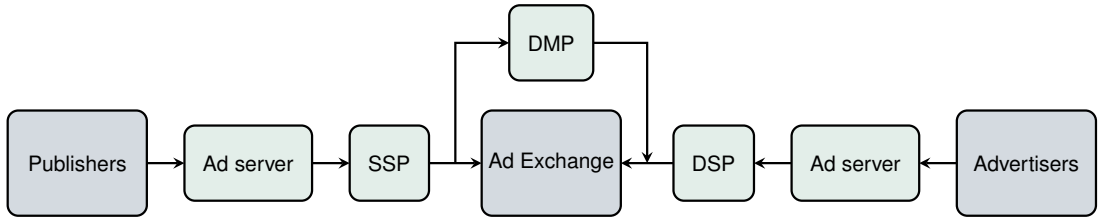
\begin{figure}
 {\centering
 \tikzset{agent/.style={minimum width=1.85cm, minimum height=1.4cm, line width=0.3mm, fill=UmUBlue!20, rounded corners, font=\sffamily\footnotesize,inner sep=1pt, outer sep=0pt}}

\tikzset{helper/.style={ minimum height=1cm,line width=0.3mm, fill=UmUGreen!20, rounded corners, font=\sffamily\footnotesize,inner sep=7pt, outer sep=0pt}}

\begin{tikzpicture}

\node [draw,
 agent
] (advertiser) at (0,0) {Publishers};
 
\node [draw,
    helper,
    right=.5cm of advertiser
]  (adserver1) {Ad server};

\node [draw,
    helper,
    right=.5cm of adserver1
]  (DSP) {SSP};

\node [draw,
    agent,
    right=.7cm of DSP
]  (adexchange) {Ad Exchange};

\node [draw,
    helper,
    right=.7cm of adexchange
]  (SSP) {DSP}; 

\node [draw,
    helper,
    above=.4cm of adexchange
]  (DMP) {DMP};

\node [draw,
    helper,
    right=.5cm of SSP
]  (adserver2) {Ad server}; 

\node [draw,
    agent,
    right=.5cm of adserver2
]  (publisher) {Advertisers};  

\draw[-stealth, line width=0.3mm] (advertiser.east) -- (adserver1.west);
\draw[-stealth, line width=0.3mm] (adserver1.east) -- (DSP.west);
\draw[-stealth, line width=0.3mm] (DSP.east) -- (adexchange.west);
\draw[-stealth, line width=0.3mm] (SSP.west) -- (adexchange.east);
\draw[-stealth, line width=0.3mm] (adserver2.west) -- (SSP.east);
\draw[-stealth, line width=0.3mm] (publisher.west) -- (adserver2.east);

\node (fork1) at ($(DSP.east)!0.5!(adexchange.west)$) {};
\node (fork2) at ($(SSP.west)!0.5!(adexchange.east)$) {};

\draw[stealth-,  line width=0.3mm] (DMP.west) -| (fork1.center);
\draw[stealth-,  line width=0.3mm] (fork2.center) |- (DMP.east);
\end{tikzpicture}
 
 }
\caption{A schematic overview of the information flow in programmatic advertising. The publisher's ad server is informed by a supply-side platform (SSP) about what ad to show to a consumer on behalf of a publisher. To decide, the SSP puts the opportunity for an impression up for auction on an ad exchange. Advertisers place their bids and eventually provide their ads via a demand-side platform (DSP). They are frequently helped by a so-called trading desk (not shown in image) and make use of data management platforms (DMPs) that offer additional data, often derived from cookies and browser data, but more recently also contextual information. The ad exchange implements the auction and propagates information about the winning advertiser and ad back to the ad server.}
\label{fig:information}
\end{figure}

Programmatic advertising takes off from the observation that every time a consumer requests some piece of content from an online publisher, there is an opportunity to display an ad. The consumer's exposure to the ad is called an \emph{impression}. Before serving the requested content, the publisher solicits bids on the impression from advertisers through a digital ad exchange. The winning advertiser provides an ad, which is embedded in the content and sent to the consumer. The time for the whole transaction is approximately~300 ms (a blink of an eye takes 300-400 ms).  
Figure~\ref{fig:information} describes the information flow in  these programmatic auctions. 
Two characteristic features are \emph{granularity} and \emph{automation}~\citep[Chapter 2]{Busch:2016}. Granularity enables advertisers to make optimal use of their budget, by inspecting each individual impression and bidding in proportion to its expected statistic outcome in terms of, for example, consumer activation. This detail-oriented view is only viable due to automation, which avoids expensive and time-consuming manual processing.  

\subsection{Classification and clustering}
There are several uses of AI in programmatic auctions. A prime example is the classification of impressions into high-level categories that have predictive value for advertising effectiveness. When such a classification is applied to consumer data, we talk about \emph{segmentation} of audiences. When instead the classification is made with respect to the media content, we have contextual advertising. It is also possible to classify consumer behaviour, for example, how long they spend on a web page, and how they navigate through the site. Finally, classification can be used to match user queries in search engines with sponsored links \citep{Chakrabarti:2008,Richardson:2007,Wang:2011}. We leave a deeper treatise of contextual search advertising as an item for future work. 

Since our focus is on contextual advertising, we note that AI-based content classification can be used to identify or at least estimate the three previously mentioned context factors: (1) the applicability of the content through \emph{topic identification}~\citep{JelodarEtAl:2019}, (2) the affective tone through \emph{sentiment analysis}~\citep{YadavEtAl:2020}, and (3) the involvement of the consumer through correlating factors such as time spent on the website, scroll pace, or similar content previously consumed~\citep{Kwon:2019}. Topic identification and semantic analysis are instances of natural language processing and usually focus on textual content, but the techniques are increasingly applied to multimodal content that also includes, e.g., images and sound~\citep{Chen:2016,Sujuan:2017}.

Topic identification is particularly useful for CA. News publishers are already tagging their articles with broad categories such as `sport', `economy', and `lifestyle' and make this metadata available to help inform bids in programmatic auctions. However, manual tagging is time consuming and it is difficult have consistency between annotators within a publication, let alone between publications. Here, topic identification can help by automating the processes and ensuring a coherent use of metadata tags. Topic identification can also be initiated by an advertising brand to detect media contexts relevant for them. For example, `sugar-free ice cream' is a too niched category to be meaningful for many publishers to tag proactively, since it is not likely that they can reliably offer updated content in the category. However, a brand advertising through a range of publishers may be able to identify sufficiently many articles matching this description to run a successful campaign. Topic identification realized through deep learning can also find abstract categories such as `relaxation' or `self fulfillment' which are difficult to reduce to individual keywords in article text. 

Another application of AI is to cluster articles or advertisements into groups with similar traits. It may for example be discovered that a certain group of articles is particularly useful to promote household services, while another is better for bank loans. Analogously, we may find that some advertisements work better in positive contexts, while others benefit from emotionally charged media. In both cases, clustering can be a good alternative to direct topic identification, as it may be computationally less demanding to sort new articles into existing clusters  than test for a vast set of fixed topics. A challenge with clustering is that the definition of similarity used to group media items must be carefully constructed, so that the resulting clusters form what we would see as a natural categories. 

\subsection{Reinforcement learning}
\label{subsec:reinforcement}
Once the auctioned impressions have been enriched with descriptive metadata through the above-mentioned techniques, one can use \emph{reinforcement learning}~\citep{sutton2018reinforcement} to automatically optimise advertising efforts with respect to different effectiveness metrics. In brief, the AI maintains a internal model of how different features of the impression contribute towards the likelihood of activating the consumer. Every time the AI wins an auction and is allowed to serve an ad, it gains a new piece of information (namely, whether the ad was clicked or not) that helps improve the model.  An advantage of reinforcement learning is that it reduces the need for \emph{a priori} knowledge about the media domain, and an AI model trained in one campaign can be reused in later campaigns for similar ads. The system may also discover valuable priming factors, e.g., that advertisements for hiking equipment perform well near articles about stress management.

An important downside of modern reinforcement learning is that  machine-learning models based on neural networks typically offer little transparency, and may come to exploit patterns in the data that are discriminatory or otherwise problematic, without the advertiser realizing. Moreover, AI systems based on reinforcement learning can be sensitive to noisy data in the early stages of the learning process, where it must base its decisions on relatively few observations. It was shown by  \cite{TolomeiEtAl:2019} that filtering out accidental clicks, which contribute to the noise, has a strong positive impact on the performance of the trained model.

As previously mentioned, the fact that programmatic advertising offers direct feedback is one of the main reasons why AI solutions are likely to work well. This is for example the case when the advertiser targets activation or conversion rates, but there are also other success criteria. For example, a brand might advertise to stay top-of-mind in a target category, or to re-position their brand, to be perceived as more environmentally friendly or exclusive. In these cases, it is difficult to develop accurate AI solutions because the results are harder to quantify and the feedback loop is longer. Sometimes the individual viewing of an ad has almost no measurable effect, and it is only by repeated exposure to the message that the consumer is affected. We therefore see it as an important item for future work to find ways of quantifying the impact of an impression on metrics such a brand recall and brand attitude. In a lab environment, this can be done by correlating data from eye tracking, emotion detection, questionnaires, etc.,  with features linked to the shown ad and media context, and the findings may then be transferable to a real-world setting.  


\section{Implications of AI-driven contextual advertising}
\label{sec:implications}

Artificial intelligence can identify suitable contexts and distribute ads efficiently. However, automated ad delivery gives rise to concerns about the fairness of the targeting mechanism. The insights gained from contextual and personal data should be used ethically, and not to target vulnerable consumers or deny others service. Furthermore, the improved capabilities for context analysis that AI brings to advertisers, also present new challenges to media publishers. Unfortunately, unfair or unethical outcomes caused by a self-optimizing AI system can be difficult to discover, let alone to predict. A deeper understanding of the implications of AI-driven advertising puts us in a better position to mitigate the risks. 

\subsection{Unethical use of context}
Contextual advertising addresses consumers when they are in a state of mind that makes them receptive towards the advertised message. This strategy can be applied unethically, by targeting disadvantaged people when their defences are down. Similarly, a self-optimizing system may discover and exploit consumer vulnerabilities revealed by their content consumption. 
Such a system may, for example, learn to place ads alongside articles about forced evictions, exploiting the media context to benefit from the reader’s emotional state. 
A particularly worrying use of manipulative advertising is within political communication. The British consultancy firm Cambridge Analytica, known for their involvement in the  2016 US presidential campaign, processed millions of people’s Facebook profiles and established sophisticated models of their personalities~\citep{cadwalladr2018revealed}. These models were used to direct personalized advertising towards undecided voters. A combination of AI-driven contextual and personal advertising could further enhance the persuasiveness of manipulative ads. Advertisers will not only know who we are, but also what media we are consuming, essentially looking over our shoulder and feeding us carefully selected messages as we go along.


\subsection{Imbalanced and discriminatory targeting of ads}

The rising popularity of contextual advertising can be seen as a reaction to the privacy and fairness issues linked to personalized advertising. 
In personalized advertising, the data used for targeting profiles mainly covers interests and behaviours, but is seldom demographic in nature. However, with the addition of AI there is a real risk that sensitive information can be derived from it, including political beliefs, sexual orientation, race or ethnicity, physical or mental health status, or sex or gender identity. Imbalanced ad delivery could constitute discrimination, if groups of consumers are offered or denied different products or prices \citep{wachter:2019}. 
In principle, contextual advertising treats consumers who request the same content equally and uses identical messaging for all visitors of a websites. However, it is likely that there are internal similarities in media consumption between members of the same group, which means that context can come to correlate with demographic groups. 
%
%

Transparency regarding how self-optimizing systems reach decisions remains one of the greatest challenges within AI, not just in advertising. Several studies on personalized advertising have confirmed discrimination based on gender~\citep{Lambrecht:2016}, race~\citep{Ali:2019}, and income level~\citep{Miller:2019}:
The delivery of ads for career opportunities in science and technology have been found to be skewed towards men \citep{Lambrecht:2016}, and  geographical targeting and dynamic pricing have caused minorities to systematically be charged higher prices \citep{Miller:2019}. 
Bias management techniques have been suggested for all parts of the machine learning process: To clean training data from discriminative patterns, to modify learning algorithms to strive for balanced representation, and to tune the trained systems to reduce biases~\citep{MehrabiEtAl:2019}. 

\subsection{Advertising stereotypes}
Stereotypical gender, race, and socioeconomic roles are staples of digital advertising. Since the ad has to capture the consumer's attention and convey its meaning within a few seconds~\citep{SajjacholapuntBall:2014}, it is tempting for the advertiser to fall back on archetypes in their storytelling. However, the societal cost is high, because advertising is an important factor in the formation of norms and cultural perspectives. The effect is amplified with contextual advertising: If an it itself fairly gender-neutral article about car repair is flanked by ads for repair tools featuring only men, then the article itself will appear to speak primarily to men. 
Other problems arise when the advertiser makes assumptions about the needs and interests of different demographic groups and target their advertising accordingly. Say, for instance, that there was a belief that women prefer pink and men blue, and that products in these colours were promoted accordingly. It should be clear that even if no such difference in tastes existed at the outset, relentless campaigning could over time come to skew the preferences of the two groups. 



\subsection{Reactive advertising}

Reactive advertising, i.e., ads that directly comment on their media context, is a particularly powerful form  of contextual advertising. The practice is enabled by AI technologies such as topic identification and semantic analysis, which provide the advertising system with a sophisticated understanding of website content, and by the programmatic ecosystem, which makes it possible to purchase ad space alongside desired content. Reactive advertising allows the advertiser to voice their stance on issues the consumer is reading about and is likely to be interested in. This is in line with the current trend of companies taking a public stand on high-profile political and social issues and movements: According to the Edelman Trust Barometer, 53\% of global consumers think that brands have a responsibility to involve themselves in at least one social issue that does not directly impact its business \citep{Edelman:2019}.

For media publishers, regulating reactive advertising could be challenging. Allowing advertisers to allude to the website content in their communication to consumers can raise the value of the ad space and the price advertisers are willing to pay.  On the other hand, an advertiser voicing controversial opinions or refuting  claims made in website content could damage the perceived integrity and credibility of the publication. We therefore expect that more reputable media publishers will ban the practice of reactive advertising altogether. Today, it is possible to block advertisers based on, e.g., IAB categories, but it will be difficult to decide in a real-time bidding context whether an ad is \emph{too} relevant to display. It is more likely that contextual advertising will be regulated through contractual means, where the advertisers agree not to use reactive advertising to comment on editorial content.  

\section{Conclusion and future work}

The media context influences how an ad is received by consumers. In contextual advertising, advertisers exploit characteristics of the medium and surrounding content to increase the effectiveness of their messaging. A favourable ad placement can improve a variety of advertising objectives, including brand perception, ad recall, click-through rate, and purchase intention. The advertising literature conjectures three context factors to be of particular importance for the consumer's perception of an ad: (1) applicability -- the topical similarity between content and ad; (2)  affective tone  -- the mood or feeling associated with the content; and (3) content involvement -- the consumer's level of engagement with the content. In an online setting, advertisers can 
use AI to leverages such factors, in order to place ads with greater precision and optimise  bids in programmatic auctions. AI is thus a main driver behind the increasing uptake of contextual advertising, and the application of AI is already extensive. However, turning over control to self-optimizing systems in the pursuit of advertising effectiveness can lead to unfair ad delivery and manipulative use of context.

 We note that much of the research on contextual advertising is dated, focusing on offline media channels including print and TV. As revealed by previous studies, the influence of different context factors varies between media channels. Contextual advertising has shown to be effective  in online display advertising \citep{Song:2014, Huang:2014}, but research is needed in other online environments such as social media, news sites, and video on demand. 
 

Another avenue for future research is to develop optimisation techniques for contextual targeting with respect to metrics such as brand perception and ad recall. Since these metrics are arguably as important as consumer activation, it would be of immense practical value if we could find ways of automatically quantifying them.  For current systems, where the main feedback on consumer interest is clicks and dwell times, this might not be possible. However, as wearable sensors become more widespread, e.g., in the form of smart watches that measure pulse and electro-dermal activity, also emotional feedback may be available. 

While some of the risks associated with AI-driven marketing are the result of intentional misuse, many of them are due to the complexity of the underlying systems, which makes the systems' behaviour difficult to analyse and predict. To provide transparency for advertisers and consumers, \cite{Rust:2020} argue that advertisers should prefer explainable AI solutions over powerful ones. 
At the date of writing it is unclear to us how the lack of transparency is best addressed, but solutions will likely have to be multidisciplinary, combining technology, self regulation, and legal means. 


\section{Implications for Advertisers}
As contextual information becomes a commodity in programmatic advertising, marketing strategies need to be updated to leverage  new opportunities and mitigate new risks.
In contrast to personalised advertising, contextual advertising provides effective means of expanding a target audience: Rather than wearing out existing customers  with repeated advertising, the marketing team can identify key context factors that reflect the desired position for their product, and use these to communicate in relevant contexts. 

Contextual advertising also enables reactive communication, where the ad is used to comment on on-going trends and events. To make the most of this opportunity, the marketing team should know in advance what kind of events would provide good talking-points for their brand, and have support from creatives that can quickly deliver custom ads for new campaigns. 
Advertisers may also want to leverage self-optimising systems, that automatically discover in what context a certain ad does well and directs ad spend accordingly. For this to be feasible, the advertiser must ensure that feedback is provided at the level of impressions rather than entire campaigns, so the system can learn by step-wise trial and error. 

For advertisers that replace personalised targeting by contextual, the consumer-privacy risks should decrease considerably. Also the risk of discriminating against different demographics should lessen, but care must be taken not to target context features that correlate strongly with, for example, the age, gender, or race of the reader.  We encourage advertisers to join one of the on-going intiatives towards self-regulation of contextual advertising hosted by, among others, the IAB and the contientous advertising network (CAN).

\newpage
\bibliographystyle{agsm}
\bibliography{references}

\end{document}